\documentclass[letterpaper]{article} 
\usepackage{aaai25}  
\usepackage{times}  
\usepackage{helvet}  
\usepackage{courier}  
\usepackage[hyphens]{url}  
\usepackage{graphicx} 
\usepackage{natbib}  
\usepackage{caption} 
\usepackage{algorithm}
\usepackage{algorithmic}

\usepackage{newfloat}
\usepackage{listings}
\DeclareCaptionStyle{ruled}{labelfont=normalfont,labelsep=colon,strut=off} 
\floatstyle{ruled}
\newfloat{listing}{tb}{lst}{}
\floatname{listing}{Listing}
\usepackage{multirow}
\usepackage{graphicx}
\usepackage{booktabs}
\usepackage{threeparttable}
\usepackage{amsmath}
\usepackage{amsfonts}
\usepackage{amssymb}  
\usepackage{colortbl}
\definecolor{mygray}{gray}{.9}
\definecolor{mypink}{rgb}{.99,.91,.95}
\definecolor{mycyan}{cmyk}{.3,0,0,0}

\title{%
  \raisebox{-0.35\height}{\includegraphics[width=1.4em]{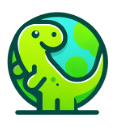}} 
  Locate Anything on Earth: Advancing Open-Vocabulary \\ Object Detection for Remote Sensing Community
}
\author{
    Jiancheng Pan\textsuperscript{\rm 1,2}\equalcontrib\thanks{
    Work is done during an internship at Tsinghua University.
    },
    Yanxing Liu\textsuperscript{\rm 3}\equalcontrib,
    Yuqian Fu\textsuperscript{\rm 4,5$\ddagger$},
    Muyuan Ma\textsuperscript{\rm 1}, \\
    Jiahao Li\textsuperscript{\rm 1},
    Danda Pani Paudel\textsuperscript{\rm 5},
    Luc Van Gool\textsuperscript{\rm 4,5},
    Xiaomeng Huang\textsuperscript{\rm 1}\thanks{Corresponding author.}
}
\affiliations{
    \textsuperscript{\rm 1}Tsinghua University\\
    \textsuperscript{\rm 2}Zhejiang University of Technology\\
    \textsuperscript{\rm 3}University of Chinese Academy of Sciences\\
    \textsuperscript{\rm 4}ETH Zürich\\
    \textsuperscript{\rm 5}INSAIT, Sofia University ``St. Kliment Ohridski"\\
    jiancheng.pan.plus@gmail.com, liuyanxing21@mails.ucas.ac.cn, yuqian.fu@insait.ai, hxm@tsinghua.edu.cn

}

\usepackage{bibentry}

\begin{document}

\maketitle

\begin{abstract}
Object detection, particularly open-vocabulary object detection, plays a crucial role in Earth sciences, such as environmental monitoring, natural disaster assessment, and land-use planning.
However, existing open-vocabulary detectors, primarily trained on natural-world images, struggle to generalize to remote sensing images due to a significant data domain gap.
Thus, this paper aims to advance the development of open-vocabulary object detection in remote sensing community. 
To achieve this, we first reformulate the task as \textbf{\textbf{L}ocate \textbf{A}nything on \textbf{E}arth (LAE)} with the goal of detecting any novel concepts on Earth.
We then developed the \textbf{LAE-Label Engine} which collects, auto-annotates, and unifies up to 10 remote sensing datasets creating the \textbf{LAE-1M} — the first large-scale remote sensing object detection dataset with broad category coverage. 
Using the LAE-1M, we further propose and train the novel \textbf{LAE-DINO Model}, the first open-vocabulary foundation object detector for the LAE task, featuring \textit{Dynamic Vocabulary Construction (DVC)} and \textit{Visual-Guided Text Prompt Learning (VisGT)} modules. DVC dynamically constructs vocabulary for each training batch, while VisGT maps visual features to semantic space, enhancing text features.
We comprehensively conduct experiments on established remote sensing benchmark DIOR, DOTAv2.0, as well as our newly introduced 80-class LAE-80C benchmark. Results demonstrate the advantages of the LAE-1M dataset and the effectiveness of the LAE-DINO method.
\end{abstract}
%
\begin{links}
    \link{Code}{https://github.com/jaychempan/LAE-DINO}
\end{links}

\section{Introduction}
As one of the most fundamental and important tasks in the file of computer vision, object detection (OD) and localization ~\cite{ren2015faster} has been extensively studied over the years, leading to the development of numerous detectors. In particular, open-vocabulary object detection (OVD)~\cite{zareian2021open} has been receiving increasing attention. OVD relaxes the limitation of close-set object categories in the traditional OD, allowing the detection of any novel concept during the testing time. Among various OVD methods, DINO~\cite{zhang2023dino} based detectors, e.g., GroundingDINO~\cite{liu2024grounding}, have recently shown promising performance on mainstream OVD benchmarks.

However, almost all of the state-of-the-art OVD methods are trained and tested on natural-world images. When applied to Earth science-related tasks, such as environmental monitoring, natural disaster assessment, land-use planning, these methods struggle to generalize due to the huge data domain gap. Unlike natural-world imagery, Earth science relies on remote sensing imagery, which exhibit much higher resolutions, distinct image styles \cite{10507076,pan2023reducing}, and different semantic class concepts. This makes the direct transfer of current OVD models nontrivial.
Therefore, in this paper, we are motivated to \textbf{\textit{advance open-vocabulary object detection for remote sensing community.}}

\begin{figure*}[t]
    \centering
    \includegraphics[width=0.65\linewidth]{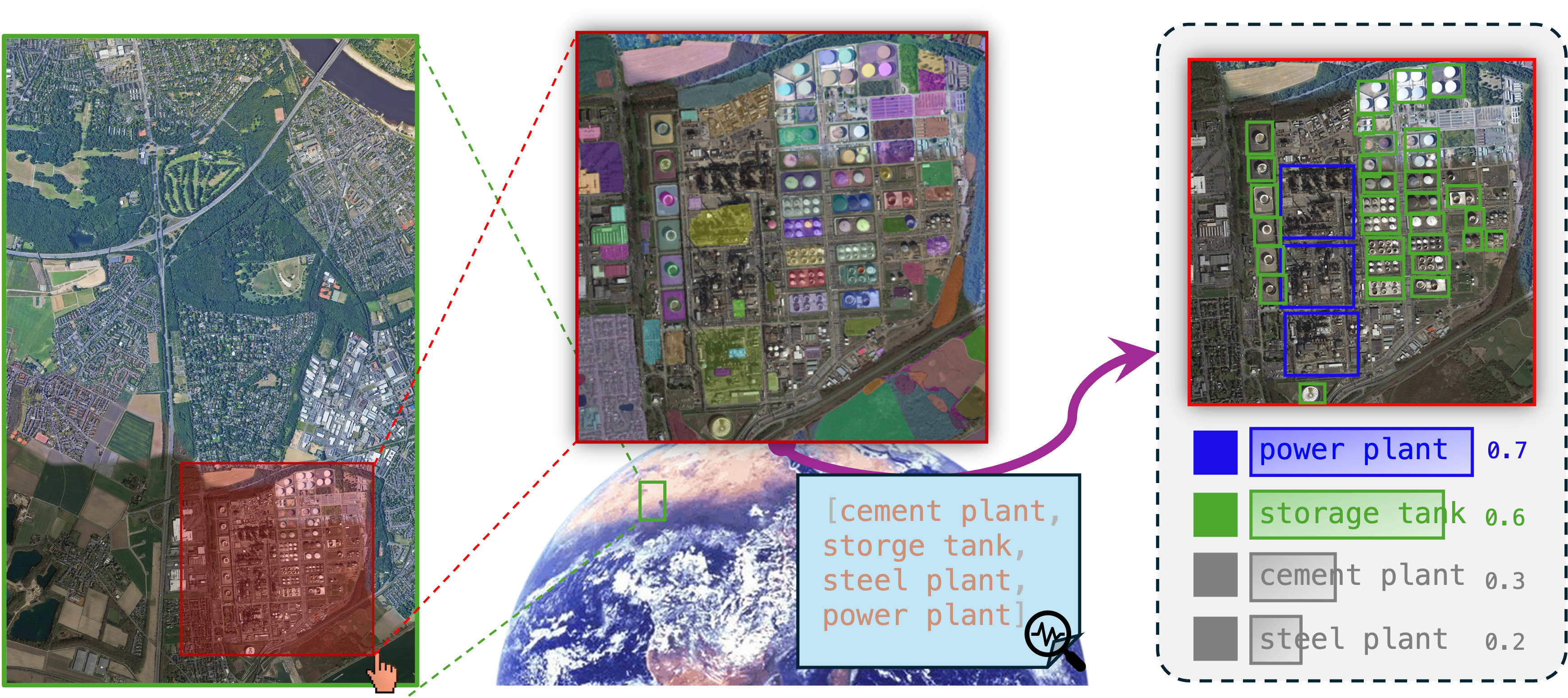}
    \caption{
    Locate Anything on Earth (LAE) aims to detect any object on Earth and facilitate practical detection tasks, powered by LAE-Label Engine and LAE-DINO Model.
    }
    \label{fig:fig1}
\end{figure*}

To achieve this goal, we first reformulate the task of OVD for remote sening filed as \textbf{\textbf{L}ocate \textbf{A}nything on \textbf{E}arth (LAE)}\footnote{
\textcolor{black}{Following \cite{zhang2024goodcaptioningbadcounting}, we use ``localization” to describe detection tasks in the remote sensing domain.}
}. As illustrated in Figure \ref{fig:fig1}, our aim is to enable LAE models could detect any novel concept on Earth. Our efforts are mainly made from two key aspects: first, a \textbf{LAE-Label Engine} is developed to construct the large-scale remote sensing training data; second, a novel \textbf{LAE-DINO Model} is proposed and trained to work as the first foundation models for the newly proposed LAE task.

More specifically, the LAE-Label engine is proposed to solve the lack of diverse object-level labeled data in the remote sensing community, which is essentially an indispensable part of training robust foundation models. 
To fully leverage the existing scattered remote sensing data which can be broadly grouped into labeled and unlabeled data, our LAE-Label engine proposes two distinct solutions.
For labeled datasets, we focus on unifying them through image slicing, format alignment, and sampling, forming the fine-grained LAE-FOD dataset. 
For unlabeled datasets, we develop a semi-automated labeling pipeline using SAM \cite{kirillov2023segment}, a large vision-language model, and rule-based filtering, resulting in the coarse-grained LAE-COD dataset.
By combining LAE-FOD and LAE-COD, we ultimately construct the \textbf{LAE-1M} dataset with one million labeled objects across diverse categories.
To our knowledge, LAE-1M is the first and largest remote sensing object detection dataset with broadest category coverage to date.

Technically, the LAE-DINO, a DINO-based OVD method, is proposed and trained on the LAE-1M dataset. The novel modules of LAE-DINO are designed to address two questions: 1) How to fit the OVD model in the training data that has around 1600 vocabularies? 2) How can the relationship between image and text be better utilized to achieve more effective vocabulary-conditioned object detection? 
As the answer of the first question, the Dynamic Vocabulary Construction (DVC) which dynamically selects the positive and negative vocabularies for each training batch is proposed. While the Visual-Guided Text Prompt Learning (VisGT) is presented to address the second issue. Based one the observation that different objects within a single image collectively define the scene, VisGT introduces the concept of ``scene features" by averaging all object features. Through taking the scene features as a bridge, VisGT aligns visual features with text features, thereby enhancing the interaction between these two modalities. Extensive experiments are conducted on both open-set and close-set scenarios. Different models are compared taking different data as training data. Results reveal: 1) our proposed LAE-1M dataset significantly improves model performance, especially in open-set scenarios; and 2) our LAE-DINO model achieves state-of-the-art performance.

We summarize the main contributions as follows,
\begin{itemize}
\item 
We advocate the Locate Anything on Earth (LAE) task for remote sensing and pave the way for LAE by contributing the LAE-1M data with one-million instances.
\item 

We propose a novel LAE-DINO detector for LAE, with dynamic vocaublary constuction (DVC) and Visual-Guided Text Prompt Learning (VisGT) as novel modules.
\item Extensive experimental results on several different testing benchmarks demonstrate the advantages of the LAE-1M dataset and the effectiveness of the LAE-DINO.
\end{itemize}

\section{Related Work}
\subsubsection{Generic Object Detection for Remote Sensing.} Object detection (OD) is one of the classical vision tasks in computer vision, which is to obtain the locations of regions of interest from a given image. Object detection methods can be mainly divided into single-stage and two-stage object detection. Single-stage methods (e.g. YOLO Family \cite{redmon2017yolo9000}) perform classification and regression directly on a predefined mass of anchor boxes. The two-stage methods (e.g. Faster R-CNN \cite{ren2015faster}) fine-tune bounding boxes based on the single-stage, which are usually more accurate than single-stage methods but are slower. Some representative works, such as DINO-based detector \cite{zhang2023dino} based on Transformer, explore the trade-off between performance and computational cost for more robust detectors. In the remote sensing community, extensive research efforts \cite{tao2023tov,cong2022satmae,reed2023scale,bastani2023satlaspretrain,sun2022ringmo,guo2024skysense} are concentrated on the extraction of fundamental imagery knowledge from large volumes of unlabeled data, utilizing advanced self-supervised or unsupervised methodologies. \textcolor{black}{Some methods (e.g. CALNet \cite{he2023multispectral}) work on involving visible (RGB) and infrared (IR) images to enhance detection performance.} While these methods are broadly applicable, they exhibit limited effectiveness in enhancing detection capabilities.

\begin{figure*}[t]
    \centering
    \includegraphics[width=0.9\linewidth]{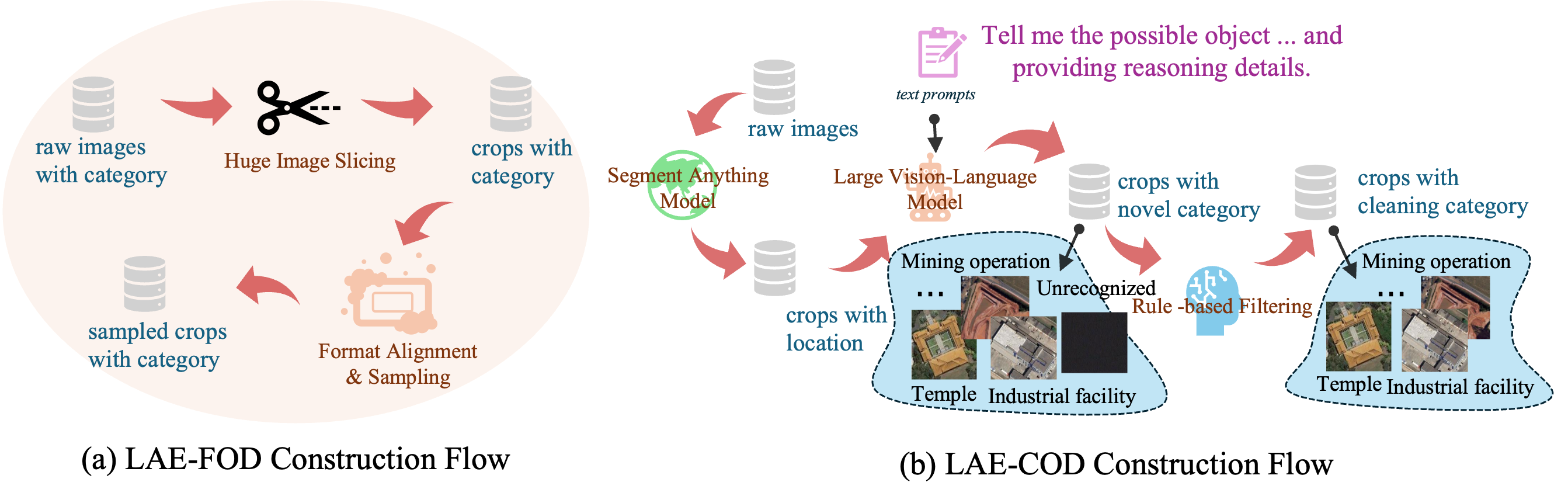}
    \caption{
    The pipeline of our LAE-Label Engine.
    }
    \label{fig:fig5}
\end{figure*}

\subsubsection{Object Detection from Few-Shot Learning to Open-Vocabulary Learning.} 
\textcolor{black}{Few-Shot Object Detection (FSOD) \cite{chen2018lstd} aims to detect unseen objects using only a few labeled examples.}
FSOD approaches are divided into fine-tuning-based \cite{chen2018lstd}, which transfers knowledge \cite{hospedales2021meta} from base to novel classes, and meta-learning-based, which uses ``learning to learn" to generalize across novel classes. 
\textcolor{black}{CD-FSOD \cite{fu2024cross}) explores cross-domain FSOD (e.g., from natural to remote sensing images), yet it relies on visual images, offering limited support for new vocabulary.}

Therefore, Open-Vocabulary Object Detection (OVD) adopts a more practice-oriented learning paradigm \cite{zareian2021open} compared with FSOD, aiming to construct an open visual-semantic space to enhance out-of-category identification and localisation. OVR-CNN \cite{zareian2021open} first proposed to acquire knowledge from natural language vocabularies by pre-training the backbone with image-caption data. After that, RegionCLIP \cite{zhong2022regionclip} and GLIP \cite{li2022grounded} became unified with the image-text matching task, expanding the visual-semantic space with more powerful flooding capabilities. While the previous work mainly improves zero-shot recognition with the help of vision-language pre-training, Grounding-DINO \cite{liu2024grounding} obtained a more robust grounding capability by introducing a stronger detector structure and fine-grained multimodal feature fusion. CasDet \cite{li2024open} combines semi-supervised learning and OVD to augment aerial detection. Due to insufficient domain annotation data, these works are weaker in open-set detection, although some show promising results in closed-set detection.

\section{Locate Anything on Earth Task}

\subsubsection{Task: Locate Anything on Earth.}
Locate Anything on Earth (LAE) draws inspiration from the Open-Vocabulary Object Detection (OVD) task but is specifically tailored for the remote sensing field. Given remote-sensing imagery as input, LAE aims to achieve robust object recognition and localization based on provided text prompts.

LAE maintains a base training dataset $\mathcal{D}_{base}$ and any potential testing dataset $\mathcal{D}_{test}$. Formally, the base dataset is represented as $\mathcal{D}_{base} = \{I, \{(b,y)_{r}\}\}$, where $I$ denotes a remote sensing image, and each image comprises $r$ objects with corresponding localization annotations $b$ and category annotations $y$. Specifically, $I$ is defined as $I \in \mathbb{R}^{H \times W \times C}$, $b$ as $b \in \mathbb{R}^{4}$, and $y$ as an element of $\mathcal{V}_{base}$, where $\mathcal{V}_{base}$ is the set of vocabularies present in $\mathcal{D}_{base}$. 
A large $\mathcal{V}_{base}$ is generally preferable for training foundational LAE models effectively. Moreover, we define $\mathcal{V}_{\Omega}$ as the entire language vocabulary and $\mathcal{V}_{test}$ as the testing vocabulary within $\mathcal{D}_{test}$. Consistent with the fundamental settings of OVD~\cite{zareian2021open}, no constraints are imposed on $\mathcal{V}_{test}$, indicating that it can be any subset of $\mathcal{V}_{\Omega}$.

Overall, LAE necessitates that models learn from $\mathcal{D}_{base}$ and subsequently identify the correct object localizations $b$ and categories $y$ for images in $\mathcal{D}_{test}$ based on the provided text prompt $\mathcal{T}$.

\subsubsection{Engine: LAE-Label Engine.}
As widely recognized, one of the essential requirements for training foundational models is the availability of large amounts of training data. Thus, naturally, this paper also aims to construct a dataset that could support the training of foundational LAE models. 
However, in the remote sensing community, the existing datasets show such limitations: 1) the human-labeled datasets are small-scale and have different sizes and data format; 2) the large-scale image-text paris which could be easily obtained from the Internet lacks well annotations.

To tackle these two limitations, we propose the LAE-Lable data engine which makes use of both the well-labeled data and the massive unlabeled data. More specifically, as shown in Figure \ref{fig:fig5}(a), for those well-labeled datasets, we first slice the huge image of different datasets and then unify the format as the same. This part results in our fine-grained LAE-FOD dataset; For the unlabeled data, as in Figure \ref{fig:fig5}(b), we build a comprehensive semi-automated data construction flow based on SAM \cite{kirillov2023segment} and Large Vision-Language Model (LVLM).

We begin by extracting the location information of Regions of Interest from remote sensing seed datasets using SAM. The detailed information of the seed datasets is listed in Table \ref{tab:table1}. Next, we obtain the categories of the zoomed-in ROI areas by taking advantage of the LVLM, i.e., InternVL \cite{chen2024internvl} which has a powerful zero-shot recognition as learned on huge amounts of data with the text prompt as demonstrated in Figure \ref{fig:fig5}(b). Finally, we filter out invalid and irrelevant categories using a rule-based method. In this way, our coarse-grained LAE-COD dataset is constructed, offering a rich vocabulary for open-vocabulary pre-training.

\begin{figure*}[t]
    \centering
    \includegraphics[width=0.9\linewidth]{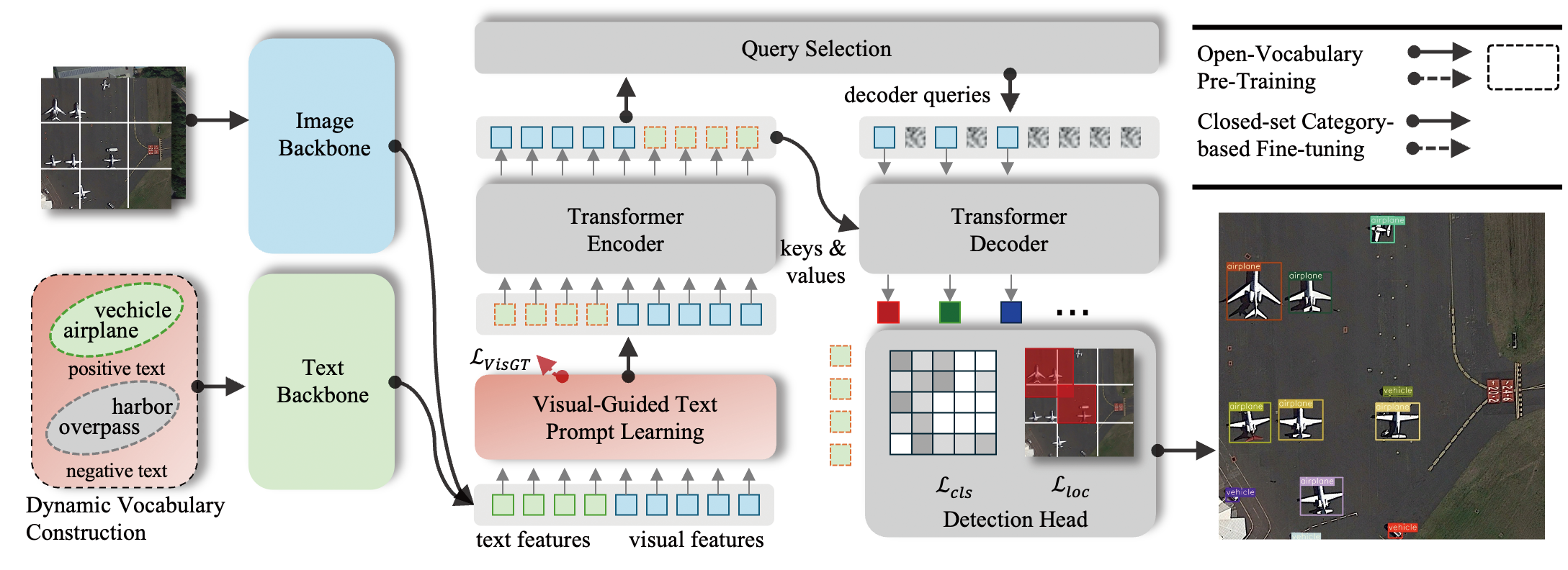}
    \caption{The pipeline for LAE-DINO.}
    \label{fig:fig2}
\end{figure*}

\section{LAE-DINO Open-Vocabulary Detector}

\subsubsection{Overview.}
Due to the huge success of DINO~\cite{zhang2023dino}, the recent DINO-based detector e.g., GroudingDINO \cite{liu2024grounding} and VideoGrounding-DINO \cite{wasim2024videogrounding}, show promising detection performance on open-vocabulary detection scenarios. Thus, in this paper, we also build our method upon the DINO and form our novel \textbf{LAE-DINO} detector.
As illustrated in Figure \ref{fig:fig2}, except for the data engine part, our LAE-DINO mainly contains the Dynamic Vocabulary Construction (DVC), the Image Backbone $E_{img}$, the Text Backbone $E_{text}$, the Visual-Guided Text Prompt Learning (VisGT), the Transformer Encoder $E_{T_E}$, the Query Selection $M_{qs}$, the Transformer Decoder $E_{T_D}$, and the Detection Head $M_{det}$. Note that the $E_{img}$, $E_{text}$, $E_{T_E}$, $M_{qs}$, $E_{T_D}$, and $M_{det}$ are basic and common modules in DINO-based detectors, thus we keep them same with the former GroudingDINO. While the DVC and the VisGT are newly proposed in this paper. Typically, the DVC is proposed to tackle the large vocabulary set issue posed by our constructed training data, and the VisGT is a novel method that uses the visual information to further guide and transform the text features. 

In the following paragraphs, we will first introduce the basic pipeline of DINO-based Detector and then present our two novel modules. 

\subsubsection{DINO-based Detectors.}
Though developed in different directions and with different new modules, the DINO-based detectors basically share the same core pipeline: Given the training dataset $\mathcal{D}_{base}$, the first thing is to construct the vocabulary set $\mathcal{V}_{base}$ by simply merging all the existing vocabularies. The vocabulary set includes positive vocabularies for categories in the images and negative words for those not seen during training.

For each batch training iteration, as indicated in Figure \ref{fig:fig2}, the image backbone $E_{img}$ and the text backbone $E_{text}$ are used to extract the visual features $F_I \in \mathbb{R}^{n_{I} \times d}$, the text features  $F_T \in \mathbb{R}^{n_{T} \times d}$ from the input image $I$ and vocabulary set $\mathcal{V}_{base}$, respectively. The $n_{I}$ and $n_{T}$ mean the number of image and text tokens, while the $d$ denotes the dimension of features. Usually, the  Swin-Transformer \cite{liu2021swin} is used as the $E_{img}$ and the BERT \cite{devlin2018bert} is used as the $E_{text}$. 
In addition, since the $\mathcal{V}_{base}$ contains both the positive and negative vocabularies, we further denote the text features generated from $n_{T_p}$ positive vocabularies as $F_T^{P} = [\tilde{F}_{T_1}, \tilde{F}_{T_2},..., \tilde{F}_{T_p}] \in \mathbb{R}^{n_{T_p} \times d}$.

After that, the Transformer encoder $E_{T_{E}}$ which takes both the image features $F_{I}$ and the text features $F_{T}$ are applied to fuse the multi-modal features. Then, the query selection $M_{qs}$ is used to initialize the region queries which consists of the learnable content queries and dynamic positional queries. 
Finally, the Transformer decoder $E_{T_{D}}$ and the detection head $M_{det}$ output the both the location and category predictions $\{(\hat{b},\hat{y})_{r}\}$ for modality alignment.

Upon the predictions $\{(\hat{b},\hat{y})_{r}\}$ and the ground truth $\{(b,y)_r\}$, two classical losses are calculated. One is the standard Cross Entropy (CE) loss $\mathcal{L}_{cls}$ \cite{li2022grounded,liu2024grounding} for evaluating the classification results between $\hat{y}$ and $y$, another is the Generalized Intersection over Union (GIoU) loss $\mathcal{L}_{loc}$ \cite{rezatofighi2019generalized}  for evaluating the locations. The detailed calculation method for $\mathcal{L}_{cls}$ and $\mathcal{L}_{loc}$ are as follows,
\begin{equation}
\mathcal{L}_{cls} = \sum_{i=1}^{r} \mathcal{L}_{CE}(\hat{y}, y),
\end{equation}
\begin{equation}
\mathcal{L}_{loc} = \lambda_{L_1} \sum_{i=1}^{r} \mathcal{L}_{L_1}(\hat{b}, b) + \lambda_{GIoU} \sum_{i=1}^{r} \mathcal{L}_{GIoU}(\hat{b}, b).
\end{equation}

\subsubsection{Dynamic Vocabulary Construction.}
\textcolor{black}{Current OVD detectors \cite{li2022grounded,liu2024grounding} like Grounding DINO rely on fixed-token-length text encoders (e.g., BERT \cite{devlin2018bert} or CLIP \cite{radford2021learning}) that concatenate all categories into a ``extremely long text", which is nontrivial for datasets with numerous categories. For example, while BERT allows a maximum of 256 tokens, our dataset includes about 1600 categories, exceeding this limit. This motivated us to develop the dynamic vocabulary construction (DVC), which reduces the number of input categories.} 

\textcolor{black}{To tackle with such a ``extremely long text", }
APE \cite{shen2024aligning} tries to blend the individual concepts of vocabularies as independent text prompts but discarding the correlation among vocabularies.
Our DVC sets a dynamic vocabulary length $N_{\mathcal{DV}}$, for each training iteration, several positive and negative vocabularies will be selected to form the $N_{\mathcal{DV}}$ vocabulary set.
\textcolor{black}{Concretely, if the number of base vocabulary $\mathcal{V}_{base}$ is larger than $N_{\mathcal{DV}}$, i.e., $\Vert \mathcal{V}_{base} \Vert > N_{\mathcal{DV}}$. In each training batch, DVC ensures the input category length is fixed at $N_{\mathcal{DV}}$. All current batch categories are considered positive categories (for example $N_{pos}$) are included in the input. The remaining ($N_{\mathcal{DV}}-N_{pos}$) part is filled by randomly sampling negative categories from the rest of the whole category set $\mathcal{V}_{base}$. DVC can effectively reduce the number of iterations for text encoder inference.}

\subsubsection{Visual-Guided Text Prompt Learning.}

OVD models primarily reply on the relationship between image and text to achieve the open-vocabulary learning. Current DINO-based detectors, including our LAE-DINO, utilize this relations through textual prompt learning. 

However, a picture paints a thousand word which means that sparse and limited categories are hard to fully represent a image. Also inspired by MQ-Det \cite{xu2024multi}, incorporating visual prompts from additional supported images with text prompts, we propose the VisGT module which aims at leveraging the visual information to further improve the semantic representation. Notably our VisGT does not utilise visual prompts like MQ-Det, but rather visual-guided text prompts to compensate for the lack of single text prompt.

Specifically, as in Figure~\ref{fig:fig2}, VisGT is not an object-level alignment but an image-level alignment that represents the overall objects of the scene, preserving the knowledge of vocabulary to fine-grained detection across different categories.

\begin{figure}[t]
    \centering
    \includegraphics[width=\linewidth]{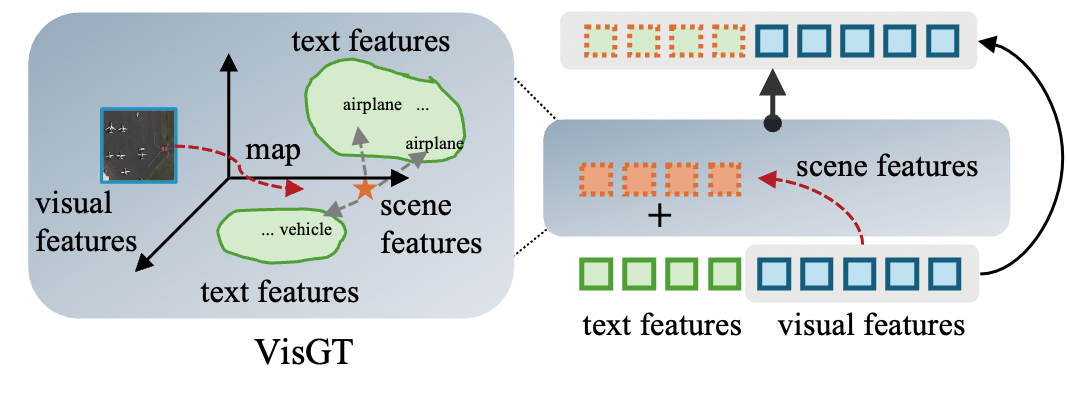}
    \caption{VisGT maps visual features into semantic space. The scene features are instance-level and category-relative features from different textual features in an image, which represents the scenographic information from the image. For example, \textit{airplane} and \textit{vehicle} belong to the \textit{airport}.}
    \label{fig:fig4}
\end{figure}

The detailed architecture of our VisGT in shown in Figure~\ref{fig:fig4}. 
First of all, we propose the ``scene features" by fusing different text features. 
The observation behind this is that the different object categories together could convert some useful scene information. 
For example, \textit{airplane} and \textit{vehicle} are two typical concepts that are strongly related to the \textit{airport} scene. Thus, given the textual features $F_{T}$ with its positive textual features $F_{T_P}$, we define the scene feature $s$ as,

\begin{equation}
s = \frac{1}{n_{T_p}} \sum_{i=1}^{T_p}  {L_i} \tilde{F}_{T_i},
\end{equation}

where $L_i$ is the token length of the $i$-th category, which corresponds to the $T_i$-th token.

By combining different instance-agnostic positive text features $F_T^{P}$, our scene features $s$ could be regarded as some special feature that contains both the instance-level and category-relative features.
This scene feature $s$ works as the ground truth when we try to map the visual information also into the semantic space.

As for the mapping of visual feature to semantic feature $\hat{s}^{l}$, we introduce the Multi-scale Deformable Self-Attention (MDSA) \cite{zhu2021deformable} as a tool as follows, 

\begin{equation}
\hat{s}^{l} = 
\left\{\begin{array}{l}
\begin{aligned}
 & \sum_{i=1}^{n_I} \frac{{F_I}_{j,:}}{n_I}, (l = 1), \\
 & FFN^l(MDSA^l(\hat{s}^{l-1})), (l > 1),
\end{aligned}
\end{array}\right.
\end{equation}
where $FFN^l(\cdot)$ is the $l$-th layer of the Feed-Forward Network, and $MDSA^l(\cdot)$ means the $l$-th of the MDSA module. 

We denote the transformed visual features as $\hat{\mathcal{S}}_{v2t}$ where the ``$v2t$" shows that our expectation of transferring the feature from visual space to textural space.

Suppose that we have already learned good $\hat{\mathcal{S}}_{v2t}$, to facilitate the enhancement of visual and textural features, we combine the original text features $F_T$ together with the $\hat{\mathcal{S}}_{v2t}$ as the input of the Transformer encoder $E_{T_E}$ as,

\begin{equation}
E_{T_E}([F_T + \hat{\mathcal{S}}_{v2t}, F_I]),
\end{equation}

\subsubsection{Constraint Loss of VisGT.}

To supervise the learning of $\hat{\mathcal{S}}_{v2t}$, we propose to use the contrastive loss \cite{hadsell2006dimensionality} as the constraint \cite{pan2023prior} between the predicted scene features $\hat{s}^l$ and predefined scene features $s$. Formally, given a batch data with $n$ images, we have the VisGT constraint loss as below,
\begin{equation}
\mathcal{L}_{VisGT} = p\left(s=\hat{s}_i^l \right)= \frac{\exp \left({\phi}_{i, i} / \tau\right)}{\sum_{j=1}^n \exp \left({\phi}_{i, j} / \tau\right)},
\end{equation}
where $\tau$ is the temperature parameter and ${\phi}_{i, j}=\hat{s}_i^{l\mathrm{T}}s_j$ denotes the similarity matrix. 

with the $\mathcal{L}_{VisGT}$ and the classical classification loss $\mathcal{L}_{cls}$, localization loss $\mathcal{L}_{loc}$, our final loss function is as,

\begin{equation}
 \mathcal{L} = \mathcal{L}_{cls} + \alpha\mathcal{L}_{loc} + \beta\mathcal{L}_{VisGT},
\end{equation}
where $\alpha$ and $\beta$ are the weight factors.

\section{Experiments}


\subsection{Experimental Setup}
\subsubsection{LAE-1M Dataset.} 

We constructed a large-scale remote sensing object detection dataset by using our  LAE-Label Engine pipeline as in Figure~\ref{fig:fig5}. As a brief recall, our dataset contains the fine-grained LAE-FOD and the coarse-grained LAE-COD. The final constructed LAE-1M dataset covered \textbf{one million} instances.
 
\begin{table}[b]
\centering
\resizebox{\linewidth}{!}{
\begin{tabular}{cc|c}
\toprule
\multicolumn{2}{c|}{\textbf{Datasets}} &  \multicolumn{1}{c}{\multirow{1}{*}{\textbf{Instances}}} \\
\hline
\multicolumn{1}{c|}{\multirow{4}{*}{LAE-COD}} & AID \cite{xia2017aid}& 34,214\\
\multicolumn{1}{c|}{} & NWPU-RESISC45 \cite{Hichri2021}& 28,906  \\
\multicolumn{1}{c|}{} & SLM \cite{9893840}& 106 \\

\multicolumn{1}{c|}{} & EMS (From Google Earth) & 39,013\\
\hline
\multicolumn{1}{c|}{\multirow{8}{*}{LAE-FOD}} & DOTA \cite{Xia_2018_CVPR}& 188,282\\
\multicolumn{1}{c|}{} & DIOR \cite{li2020object}& 192,472\\
\multicolumn{1}{c|}{} & FAIR1M \cite{sun2022fair1m}& 1.02   M \\
\multicolumn{1}{c|}{} & NWPU VHR-10 \cite{cheng2014multi}& 3,651 \\
\multicolumn{1}{c|}{} & RSOD \cite{long2017accurate}& 6,950\\
\multicolumn{1}{c|}{} & Xview \cite{lam2018xview}& $\sim$ 1 M \\
\multicolumn{1}{c|}{} & HRSC2016 \cite{liu2017high}& 2,976\\
\multicolumn{1}{c|}{} & Condensing-Tower \cite{zhang2019deep}& 2,382\\
\bottomrule
\end{tabular}}
\caption{LAE-1M dataset contains abundance categories composed of coarse-grained LAE-COD and fine-grained LAE-FOD.}
\label{tab:table1}
\end{table}



\begin{table*}[t]
\tiny
\centering
\resizebox{0.73\linewidth}{!}{%
\begin{tabular}{l|c|c|c|c}
\toprule
\multirow{2}{*}{\textbf{Method}} & \multirow{2}{*}{\textbf{Pre-Training Data}} & \multicolumn{1}{c|}{\textbf{DIOR}} & \multicolumn{1}{c|}{\textbf{DOTAv2.0}} & \multicolumn{1}{c}{\textbf{LAE-80C}} \\
 &  & $AP_{50}$ & $mAP$  & $mAP$ \\ \hline
\multicolumn{1}{l|}{GLIP \cite{li2022grounded}} & O365,GoldG,CC3M,SBU &  1.1 & 0.2  & 0.1  \\
\multicolumn{1}{l|}{GLIP with \textit{DVC} \cite{li2022grounded}} & LAE-1M & 82.8  & 43.0  & 16.5   \\
\multicolumn{1}{l|}{GroundingDINO \cite{li2022grounded}} & O365,GoldG,Cap4M & 0.3  & 0.3 & 0.1  \\
\multicolumn{1}{l|}{GroundingDINO with \textit{DVC} \cite{li2022grounded}} & LAE-1M & 83.6  & 46.0 & 17.7  \\ 
\rowcolor{mygray}\multicolumn{1}{l|}{\textbf{LAE-DINO} (Ours)} & LAE-1M & 85.5 & 46.8  & 20.2   \\
\bottomrule
\end{tabular}}
\caption{The open-set detection results on DIOR, DOTAv2.0 and LAE-80C benchmarks. All models in the table are based on Swin-T and BERT backbones. O365, GoldG, CC3M, SBU and Cap4M are natural scene datasets.
}
\label{tab:table2}
\end{table*}

\begin{table*}[t]
\tiny
\centering
\resizebox{0.8\linewidth}{!}{%
\begin{tabular}{lcccc}
\toprule
\multicolumn{1}{l|}{\multirow{2}{*}{\textbf{Method}}} & \multicolumn{1}{c|}{\multirow{2}{*}{\textbf{Backbone}}} & \multicolumn{1}{c|}{\multirow{2}{*}{\textbf{Pre-Training Data}}} & \multicolumn{2}{c}{\textbf{Fine-Tuning}} \\
\multicolumn{1}{l|}{} & \multicolumn{1}{c|}{} & \multicolumn{1}{c|}{} & DIOR($AP_{50}$) & DOTAv2.0($mAP$)  \\ \hline
\multicolumn{5}{l}{\textit{Generic   Object Detection}} \\ \hline
\multicolumn{1}{l|}{GASSL \cite{ayush2021geography}} & \multicolumn{1}{c|}{ResNet-50} & \multicolumn{1}{c|}{-} & 67.40 & - \\
\multicolumn{1}{l|}{CACO \cite{mall2023change}} & \multicolumn{1}{c|}{ResNet-50} & \multicolumn{1}{c|}{Sentinel-2} & 66.91 & - \\
\multicolumn{1}{l|}{TOV \cite{tao2023tov}} & \multicolumn{1}{c|}{ResNet-50} & \multicolumn{1}{c|}{TOV-NI,TOV-R} & 70.16 & - \\
\multicolumn{1}{l|}{Scale-MAE \cite{reed2023scale}} & \multicolumn{1}{c|}{ViT-L} & \multicolumn{1}{c|}{FMoW} & 73.81 & - \\
\multicolumn{1}{l|}{SatLas \cite{bastani2023satlaspretrain}} & \multicolumn{1}{c|}{Swin-B} & \multicolumn{1}{c|}{ SatlasPretrain } & 74.10 & -  \\
\multicolumn{1}{l|}{RingMo \cite{sun2022ringmo}} & \multicolumn{1}{c|}{Swin-B} & \multicolumn{1}{c|}{RingMoPretrain} & 75.90 & -  \\
\multicolumn{1}{l|}{SkySense \cite{guo2024skysense}} & \multicolumn{1}{c|}{Swin-H} & \multicolumn{1}{c|}{multi-modal RSI} & 78.73 & -  \\
\multicolumn{1}{l|}{MTP \cite{wang2024mtp}} & \multicolumn{1}{c|}{Swin-H} & \multicolumn{1}{c|}{MillionAID} & 81.10 & -  \\
\hline
\multicolumn{5}{l}{\textit{Open-Vocabulary   Object Detection}} \\ \hline
\multicolumn{1}{l|}{GLIP-FT \cite{li2022grounded}} & \multicolumn{1}{c|}{Swin-T} & \multicolumn{1}{c|}{O365,GoldG,CC3M,SBU} & 87.8 & 50.6  \\
\multicolumn{1}{l|}{GroudingDINO-FT \cite{liu2024grounding}} & \multicolumn{1}{c|}{Swin-T} & \multicolumn{1}{c|}{O365,GoldG,Cap4M} & 90.4 & 54.0  \\
\multicolumn{1}{l|}{GroudingDINO-FT \cite{liu2024grounding}} & \multicolumn{1}{c|}{Swin-T} & \multicolumn{1}{c|}{LAE-1M} & 91.1 & 55.1 \\
\rowcolor{mygray}\multicolumn{1}{l|}{\textbf{LAE-DINO-FT} (Ours)} & \multicolumn{1}{c|}{Swin-T} & \multicolumn{1}{c|}{O365,GoldG,Cap4M} & 92.0 & 55.5 \\
\rowcolor{mygray}\multicolumn{1}{l|}{\textbf{LAE-DINO-FT} (Ours)} & \multicolumn{1}{c|}{Swin-T} & \multicolumn{1}{c|}{LAE-1M} & 92.2 & 57.9  \\
\bottomrule
\end{tabular}}
\caption{The closed-set detection results on on DIOR and DOTAv2.0 test set.
 The results of DOTAv2.0 are all based on horizontal detection boxes. GeoImageNet, Sentinel-2, TOV-NI,TOV-R, FMoW, SatlasPretrain, MillionAID, RingMoPretrain and multi-modal RSI are remote sensing datasets.}
 \label{tab:table3}
\end{table*}

Table~\ref{tab:table1} summarizes the sub-datasets used for building the LAE-1M dataset. Specifically, for most of the datasets, a 0.4 random sampling rate is adopted if the number of instance of same class us larger than 100. Xview is the only exception, for which we sample 0.2 to eliminate the duplicate instances. The purpose of sampling instances from different classes across all datasets is to maximize the learning of each class's features while preserving the original dataset's data distribution.

\subsubsection{Evaluation Benchmarks.}
To evaluate the validity of our LAE-1M dataset and LAE-DINO model,  DIOR \cite{li2020object} and DOTAv2.0 \cite{Xia_2018_CVPR} which are commonly used in the remote sensing community are used as benchmarks as in MTP\cite{wang2024mtp}. 
Note that the results of DOTAv2.0 are all based on horizontal detection boxes for building a foundational location detector. In addition, to better validate the open-set detectors, we constructed LAE-80C containing 80 classes as a new remote sensing OVD benchmark. More details are included in Appendix. Based on the above three benchmarks, both the open-set and closed-set detection capabilities are evaluated. Specifically, we introduce the HRRSD \cite{zhang2019hierarchical} dataset with a total of thirteen classes, which contains ten base classes appearing in LAE-1M dataset and three novel classes that do not, to perform the few-shot detection experiments.The $mAP$, $AP_{50}$, and $AP_{75}$ are used as the evaluation metrics.

\subsubsection{Implementation Details.}

We conducted all pre-training experiments on four A100 GPUs. To avoid memory overflow caused by having too many objects in a single image during batch training, we split image annotations with over 200 objects into smaller groups, ensuring the number of instances remains unchanged. Additionally, the alignment heads' categories are set to 1600 for open-vocabulary pre-training. 
During training, key parameters are carefully set: the length of the dynamic vocabulary number in DVC module i.e., $N_{\mathcal{DV}}$ is set to 60, the number of layers $l$ for MDSA and FFN is set to 7, and the hyper-parameters $\alpha$ and $\beta$ of the loss function are set to 1 and 10, respectively. 
The open-vocabulary pre-training of LAE-DINO lasts approximately 180$K$ steps, spanning about 48 GPU hours with a batch size of 2 per GPU. More details are provided in \textit{More Implementation Details} section in Appendix.

\subsection{Detection Results}
\subsubsection{Open-Set Detection.}
We compare the open-set detection results with two effective OVD methods, GLIP \cite{li2022grounded} and GroudingDINO \cite{liu2024grounding}, trained on natural and remote sensing scenes datasets as shown in Table \ref{tab:table2}. 
\textcolor{black}{Note that due to the natural difference of tasks, those CLIP-based \cite{wang2024skyscript} and grounding \cite{mall2024remote,kuckreja2024geochat,li2024language} methods are not considered as competitors}. To train on LAE-1M dataset, we similarly introduce DVC on the GLIP and GroudingDINO. First of all, the detection results find that the OVD method of pre-training on natural scene dataset hardly works on remote sensing open-set detection, indicating a substantial gap between remote sensing and natural scene. Secondly, GroudingDINO has a more powerful open-set detection capability compared to GLIP from DIOR and LAE-80C. Clearly, our LAE-DINO has a better open-set detection compared with GroudingDINO, with an increases of 1.9\%, 0.8\%, and 2.5\% on DIOR, DOTAv2.0, and LAE-80C benchmarks, respectively. These detection results show that our LAE-DINO has a more robust open-set detection in the remote sensing field. 

\subsubsection{Closed-Set Detection.}
To prove the benefits of OVD, we perform fine-tuning experiments in remote sensing scenes, comparing some generic detectors (GD) on DIOR and DOTAv2.0 datasets as shown in Table \ref{tab:table3}. Most previous GDs are fine-tuned to object detection datasets after pre-training on remote-sensing images using a self-supervised approach. We directly cite the original paper results due to the lack of open source for these generic detectors. For the OVD methods, we also provide results of fine-tuning experiments based on pre-training on natural scene datasets. Comparing the GD and OVD methods, it shows that the OVD method, which introduces textual prompts, is significantly higher than the GD method with a raise of about 6\% at least in the $AP_{50}$ on DIOR's closed-set detection. The results of LAE-DINO fine-tuned on DIOR demonstrates that an outstanding performance on the DOTAv2.0, with a $mAP$ of 57.9, an increase of 2.8\% compared with GroundingDINO.

Table \ref{tab:table6} shows that the closed-set detection results on DIOR test set with the fine-tuning data randomly sampled at different scales DIOR-\textit{full}, DIOR-½, and DIOR-¼ from the DIOR train set. We find that with just half of the DIOR train set, the $AP_{50}$ could reach 89.1. This detection results shows that only a small amount of data is needed to fine-tune after open-vocabulary pre-training, which can achieve satisfactory results in real-world detection tasks.

\begin{table}[h]
\tiny
\centering
\resizebox{0.70\columnwidth}{!}{%
\begin{tabular}{l|c|c}
\toprule
\multirow{2}{*}{\textbf{Method}} & \multirow{2}{*}{\textbf{Fine-Tuning Data}} & \multicolumn{1}{c}{\textbf{DIOR}} \\ &  & $AP_{50}$ \\ \hline
LAE-DINO  & DIOR-\textit{full} & 92.2  \\
\rowcolor{mygray}LAE-DINO   & DIOR-½ & 89.1  \\
LAE-DINO & DIOR-¼ & 85.6 \\
\bottomrule
\end{tabular}}
\caption{The closed-set detection results on the DIOR test set with the fine-tuning data randomly sampled at different scales from the DIOR train set.}
\label{tab:table6}
\end{table}

\subsection{Ablation Studies}

\subsubsection{VisGT Analysis.} We perform ablation experiments on DIOR test set to explore the specific role of VisGT as shown in Table \ref{tab:table4}. \textit{LAE-1M Pre-Training} is the open-set detection results, and \textit{DIOR Fine-Tuning} is closed-set detection results that are directly fine-tuned on DIOR training dataset. 

\begin{table}[h]
\tiny
\centering
\resizebox{0.7\linewidth}{!}{%
\begin{tabular}{l|c|c}
\toprule
\multirow{2}{*}{\textbf{Method}} & \multirow{2}{*}{\textbf{Pre-Training Data}} & \multicolumn{1}{c}{\textbf{DIOR}} \\ &  & $AP_{50}$ \\ \hline
\multicolumn{3}{l}{\textit{LAE-1M Pre-Training}} \\ \hline
 $PT\text{-}baseline$ & LAE-1M & 83.6 \\
\rowcolor{mygray}$+$ VisGT  & LAE-1M & 85.5  \\
\hline
\multicolumn{3}{l}{\textit{DIOR Fine-Tuning}} \\ \hline
$FT\text{-}baseline$ & - & 89.9 \\
\rowcolor{mygray}$+$ VisGT  & - &  92.0  \\
\rowcolor{mygray}$+$ VisGT  & LAE-1M  &  92.2  \\ \bottomrule
\end{tabular}}
\caption{The ablation results on the DIOR test set. \textcolor{black}{PT-baseline denotes the Pre-Training baseline, and FT-baseline denotes the Fine-Tuning baseline.}}
\label{tab:table4}
\end{table}

From the \textit{LAE-1M Pre-Training} experiment, the group with VisGT achieved a 1.9\% increase in $AP_{50}$ for DIOR's open-set detection. This result indicates that our VisGT enhances the understanding of complex remote sensing scenes by incorporating visual-guided text prompts. We also found a further improvement after \textit{DIOR fine-tuning}, with an increase to 92.2 at $AP_{50}$, and further support for VisGT.

\subsubsection{LAE-1M Analysis.} To explore how LAE-COD and LAE-FOD of LAE-1M work, we set up two sets of comparison experiments on our LAE-DINO as shown in Appendix. We find that the detection of base classes in LAE-FOD can be improved by adding additional LAE-COD for pre-training, where the $mAP$ of DOTAv2.0 test set can be improved by 2.3\%. This also implies the feasibility of our LAE-Label to help interpret common categories of remote sensing imagery. As for the annotation quality and novel class detection of the LAE-Label engine, its survey report is in Appendix.

\subsection{VisGT Reanalysis \& Visualisation}
We set different weights to $\mathcal{L}_{VisGT}$ to observe its impact on detection performance. The reanalysis of VisGT and the visualisation of detection results are in Appendix.

\section{Conclusion}
In this paper, we introduced the Locate Anything on Earth (LAE) task, focusing on achieving open-vocabulary object detection for remote sensing. To advance the development of LAE, we concentrated on two key areas: 1) \textbf{Data:} We developed the LAE-Label Engine, a semi-automated labeling pipeline that collects and annotates data from up to 10 datasets. Using the LAE-Label Engine, we constructed LAE-1M, the first large-scale remote sensing object detection dataset. 2) \textbf{Model:} We presented LAE-DINO, a foundational open-vocabulary object detector for the LAE task, validated for its robust and generalizable detection capabilities.
We believe our work will greatly advance Earth science applications by defining a clear task, providing large-scale training data, and offering a foundation model.

\section{Acknowledgments}
This work was supported by the National Natural Science Foundation of China (42125503, 42430602). This work also was partially funded by the Ministry of Education and Science of Bulgaria (support for INSAIT, part of the Bulgarian National Roadmap for Research Infrastructure).

\bibliography{aaai25}
\clearpage
\twocolumn[
    \begin{center}
        \LARGE \textbf{Supplementary Material for Locate Anything on Earth: Advancing Open-Vocabulary Object Detection for Remote Sensing Community}
         \vspace{2\baselineskip}  
    \end{center}
]
\section{More Related Work}
\subsubsection{Data Engine powered by Large Model.}

Large models such as CLIP \cite{radford2021learning} and GPT \cite{radford2018improving} are empowered to change the current paradigm by the \textbf{Emergent Ability} due to the massive amounts of data fed to them for training. In the remote sensing community, multimodal work including a range of CLIP-related work \cite{liu2024remoteclip,zhang2024rs5m,wang2024skyscript,pan2024pir} has also emerged. Then came a bunch of vision-related large models with data-driven large-scale training, e.g., SAM \cite{kirillov2023segment}, InternVL \cite{chen2024internvl}, etc. Due to the tremendous zero-shot identification ability of the models, some works are based on these models as data engines for automated data labeling. \citeauthor{SAMRS} leverage SAM and existing remote sensing object detection datasets to build a pipeline for generating a large-scale remote sensing segmentation dataset SAMRS \cite{SAMRS}. \citeauthor{HUANG2024103061} explores the potential application of SAM to process medical images for fine-grained segmentation. As the accuracy requirements of the segmentation task are high, if it is used directly for labeling without adding human checks, it may still be limited in real segmentation.

SAM can sense the exact object edge under the specified point or box prompts but is unaware of the specific category. Large Vision-Language Models (LVLMs), such as CLIP, InternVL, etc., can recognize the relationship between images and text because of their alignment training in large-scale web image-text pair data. Although some work has attempted to give SAM the ability to perceive categories \cite{chen2023semantic}, this ability is not mature enough for actual segmentation in remote sensing, and the segmentation quality is degraded. Combined with SAM and LVLMs capabilities, good quality raw labeled data can be obtained.

\subsubsection{Prompt-based Object Detection.} 
Unlike traditional closed-set detection, open-set object detection is a significant change, allowing detector to recognize objects beyond a fixed set of categories with prompt learning. Prompt-based object detection methods can be classified into text-prompted and visual-prompted object detection.

Text-prompted object detection \cite{gu2021open,li2022grounded,liu2024grounding} focuses on guiding visual feature representations by encoding text features that enable them to find the location of regions given textual prompts accurately. This approach often uses a BERT \cite{devlin2018bert} or CLIP \cite{radford2021learning} text encoder as the text backbone. Visual-prompted object detection \cite{xu2024multi,jiang2024t} mainly defines a visual template as a prompt from the supporting image set to mine the position information of the visual objects. MQ-Det \cite{xu2024multi} incorporates visual prompts from additional supported images with text prompts. However, practical detection is more challenging in obtaining this visual template in advance. We propose the VisGT based on visual-guided text prompt learning, which aims to leverage the visual information further to improve the semantic representation and compensate for the lack of a single text prompt.

\subsubsection{Dynamic Vocabulary Strategy.} In natural language processing tasks, the over of the target vocabulary to be processed will not only affect the training speed but also the quality of sentence generation. \citeauthor{wu2018neural} propose a dynamic vocabulary sequence-tosequence (DVS2S) model which allows each input to possess vocabulary in decoding. \citeauthor{lakew2018transfer} propose a method to transfer knowledge across neural machine translation (NMT) models by means of a shared dynamic vocabulary. Unlike these approaches, our dynamic vocabulary strategy reduces the training vocabulary by randomly selection of samples to address the practical problems in open-vocabulary object detection.

\begin{figure}[t]
    \centering
    \includegraphics[width=\linewidth]{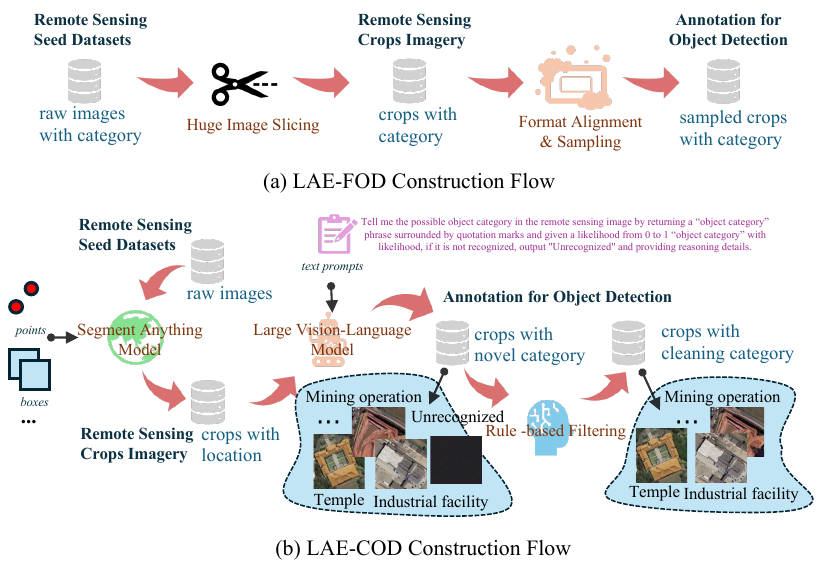}
    \caption{
    The pipeline of our LAE-Label Engine.
    }
    \label{fig:figA5}
\end{figure}

\begin{table*}[t]
\tiny
\centering
\resizebox{0.8\linewidth}{!}{
\begin{tabular}{cc|cc|cc|c|c}
\toprule
\multicolumn{2}{c|}{\multirow{2}{*}{\textbf{Datasets}}} & \multicolumn{2}{c|}{\textbf{Image Width}} & \multicolumn{2}{c|}{\textbf{Images}} & \multicolumn{1}{c|}{\multirow{2}{*}{\textbf{Instances}}} & \multirow{2}{*}{\textbf{Categories}} \\
\multicolumn{2}{c|}{} & $ori.$ & \multicolumn{1}{c|}{$pre.$} & $ori.$ & \multicolumn{1}{c|}{$pre.$} & \multicolumn{1}{c|}{} &  \\
\midrule
\multicolumn{1}{c|}{\multirow{4}{*}{LAE-COD}} & AID \cite{xia2017aid}& 600 & - & 10,000 & - & 34,214 &  1,380\\

\multicolumn{1}{c|}{} & NWPU-RESISC45 \cite{Hichri2021}& 256 & - & 31,500 & - & 28,906 & 1598  \\
\multicolumn{1}{c|}{} & SLM \cite{9893840}& 3,000$\sim$10,001 & 1,024 & 22 & 152 & 106 & 1,081 \\

\multicolumn{1}{c|}{} & EMS (From Google Earth) & 4,864$\sim$11,520 & 1,024 & 102 & 2,605 & 39,013 & 1,502 \\
\midrule
\multicolumn{1}{c|}{\multirow{8}{*}{LAE-FOD}} & DOTA \cite{Xia_2018_CVPR}& 800$\sim$4,000 & 1,024 & 2,806 & 17,480 & 188,282 & 18 \\
\multicolumn{1}{c|}{} & DIOR \cite{li2020object}& 800 & - & 23,463 & - & 192,472 & 20 \\
\multicolumn{1}{c|}{} & FAIR1M \cite{sun2022fair1m}& 1,000$\sim$10,000 & 600 & 15,266 & 64,147 & 1.02   M & 5(37) \\
\multicolumn{1}{c|}{} & NWPU VHR-10 \cite{cheng2014multi}& 533$\sim$1,728 & - & 800 & - & 3,651 & 10 \\
\multicolumn{1}{c|}{} & RSOD \cite{long2017accurate}& 512$\sim$1,961 & - & 976 & - & 6,950 & 4 \\
\multicolumn{1}{c|}{} & Xview \cite{lam2018xview}& 2,576$\sim$5,121 & 1,024 & 1,129 & 26,543 & $\sim$   1 M & 60 \\
\multicolumn{1}{c|}{} & HRSC2016 \cite{liu2017high}& 300$\sim$1,500 & - & 1,061 & - & 2,976 & 1 \\
\multicolumn{1}{c|}{} & Condensing-Tower \cite{zhang2019deep}& 304$\sim$1481 & - & 892 & - & 2,382 & 4 \\
\bottomrule
\end{tabular}}
\caption{LAE-1M Dataset is composed of coarse-grained LAE-COD dataset and fine-grained LAE-FOD dataset. "-" indicates present ($pre.$) agreement with original ($ori.$). LAE-1M does not count instances of overlap duplicates when slicing.}
\label{tab:tableA1}
\end{table*}

\begin{table*}[t]
\centering
\resizebox{\linewidth}{!}{
\begin{tabular}{c|c|c|c|c|c|c|c|c|c|c|c|c|c}
\toprule
\textbf{id} & \textbf{area} & \textbf{bbox\_x0} & \textbf{bbox\_y0} & \textbf{bbox\_w} & \textbf{bbox\_h} & \textbf{point\_input\_x} & \textbf{point\_input\_y} & \textbf{predicted\_iou} & \textbf{stability\_score} & \textbf{crop\_box\_x0} & \textbf{crop\_box\_y0} & \textbf{crop\_box\_w} & \textbf{crop\_box\_h} \\ \midrule
0  & 1939 & 161      & 210      & 62     & 45     & 197.21875      & 229.28125      & 0.9753        & 0.9574           & 85           & 85           & 171         & 171         \\ 
1  & 593  & 197      & 164      & 35     & 22     & 197.21875      & 175.84375      & 0.9736        & 0.9715           & 85           & 85           & 171         & 171         \\ 
2  & 1621 & 216      & 208      & 39     & 47     & 229.28125      & 229.28125      & 0.9694        & 0.9639           & 85           & 85           & 171         & 171         \\ 
3  & 153  & 162      & 241      & 14     & 14     & 165.15625      & 250.65625      & 0.9659        & 0.9935           & 85           & 85           & 171         & 171         \\ 
4  & 60   & 185      & 135      & 6      & 8      & 186.53125      & 143.78125      & 0.9605        & 1.0000           & 85           & 85           & 171         & 171         \\ 
5  & 158  & 127      & 248      & 25     & 7      & 143.78125      & 250.65625      & 0.9576        & 0.9568           & 85           & 85           & 171         & 171         \\
6  & 44   & 208      & 120      & 5      & 7      & 207.90625      & 122.40625      & 0.9563        & 0.9778           & 85           & 85           & 171         & 171         \\ \bottomrule
\end{tabular}}
\caption{The sample of SAM part for obtaining the RoI of an image.}
\label{tab:tableA4}
\end{table*}

\begin{table*}[h]
\centering
\resizebox{\linewidth}{!}{
\begin{tabular}{c|p{10cm}|c|c}
\toprule
\textbf{det\_name} & \textbf{text} & \textbf{class} & \textbf{likelihood} \\ \midrule
cropped\_resized\_image0\_192.0\_0.0\_589.0\_79.0\_390.5\_39.5.jpg & "Road" with a likelihood of 0.9. The image shows a paved surface with clear lane markings, which is characteristic of a road. The likelihood is not 1 because the image is not clear enough to provide a definitive identification, and there could be other similar-looking features that are not roads. & Road & 0.9 \\
cropped\_resized\_image112\_0.0\_338.0\_599.0\_479.0\_299.5\_408.5.jpg & "Airport runway" with a likelihood of 1. The image shows a large, flat, and elongated surface with distinct markings and lines, which are characteristic of an airport runway. The presence of multiple airplanes parked or in motion further supports the identification of this as an airport runway. & Airport runway & -1 \\ 
cropped\_resized\_image113\_0.0\_0.0\_599.0\_247.0\_299.5\_123.5.jpg & "Airport" with a likelihood of 0.9. The image shows a complex network of runways, taxiways, and terminal buildings, which are characteristic features of an airport. The presence of multiple airplanes parked at gates further supports the categorization of this image as an airport. & Airport & 0.9 \\
cropped\_resized\_image118\_501.0\_440.0\_599.0\_599.0\_550.0\_519.5.jpg & "Runway" with a likelihood of 0.9. The image shows a large, flat, and elongated area with distinct markings and a pattern that is characteristic of an airport runway. The markings and the way the area is surrounded by other airport infrastructure support this classification. The likelihood is not 1 because the image is not of high resolution, and some details may not be clearly visible. & Runway & 0.9 \\
cropped\_resized\_image119\_9.0\_298.0\_247.0\_359.0\_128.0\_328.5.jpg & "Airplanes" with a likelihood of 0.9. The image shows a row of objects that resemble the tail sections of airplanes, which are typically found at airports. The presence of multiple tails in a linear arrangement, along with the context of an airport setting, strongly suggests that these are indeed airplanes. The likelihood is not 1 because the image is not clear enough to confirm the details of the airplanes, such as their make or model, but the overall shape and arrangement strongly indicate that they are airplanes. & Airplanes & 0.9 \\
cropped\_resized\_image122\_215.0\_509.0\_356.0\_599.0\_285.5\_554.0.jpg & "Unrecognized" with a likelihood of 0.8. The image is too blurry and lacks clear features to confidently determine the object category. The low resolution and lack of distinct shapes or patterns make it impossible to accurately identify the object. & Unrecognized & 0.8 \\ \bottomrule
\end{tabular}}
\caption{The sample of LVLM part for annotating the RoI of an image.}
\label{tab:tableA5}
\end{table*}

\section{More Dataset Details}
\subsubsection{LAE-Label Engine.}
LAE-Label Engine consists of the LAE-FOD dataset construction and the LAE-COD dataset construction. 1) \textbf{For LAE-FOD dataset construction}, we first standardize the dataset annotation format to match the COCO \cite{lin2014microsoft} format. Then, as there are resolution oversized images in some datasets, e.g., DOTA \cite{Xia_2018_CVPR}, FAIR1M \cite{sun2022fair1m} and Xview \cite{lam2018xview}, we process to slice the images, as shown in Table \ref{tab:tableA1}. In particular, we use the sliced images and annotations from MTP \cite{wang2024mtp}, while Xview uses the open-source tool SAHI \cite{akyon2022sahi} for slicing to 1024 size. 2) \textbf{For LAE-COD dataset construction}, we first utilize two large-scale datasets for scene classification, AID \cite{xia2017aid} and NWPU- RESIC45 \cite{Hichri2021} with a low resolution images, to expand remote sensing scenarios. We then take high-resolution imagery from the constituent EMS and SLM \cite{9893840} datasets from Google Earth\footnote{https://earth.google.com/} for cropping. Finally, these raw images are fed to the LAE-COD annotation procedure mentioned in Figure \ref{fig:figA5}. We can also divide it into three parts as follows: a) \textbf{SAM Part}: We first get the region of interest (RoI) by randomly picking points as prompts and then filter the top $K$ objects with the most largest area to be saved as shown in Table \ref{tab:tableA4}. b) \textbf{LVLM Part}: We use weights from the InternVL-1.5 version, which are then fed to a specific prompt template ``\textit{Tell me the possible object category in the remote sensing image by returning a ``object category" phrase surrounded by quotation marks and given a likelihood from 0 to 1 ``object category" with likelihood, if it is not recognized, output ``Unrecognized" and providing reasoning details.
}", which then yields the text for each RoI as shown in Table \ref{tab:tableA5}. Empirically, we have found that output categories are more accurate if the LVLM provides the reason for the inference. c) \textbf{Rlue Part}: We first remove some cropped monotone images and categories for ``\textit{Unrecognized}", and some low likelihood prediction categories. Then, some culling of less accurate categories will be performed.

\begin{figure}[h]
    \centering
    \includegraphics[width=0.9\linewidth]{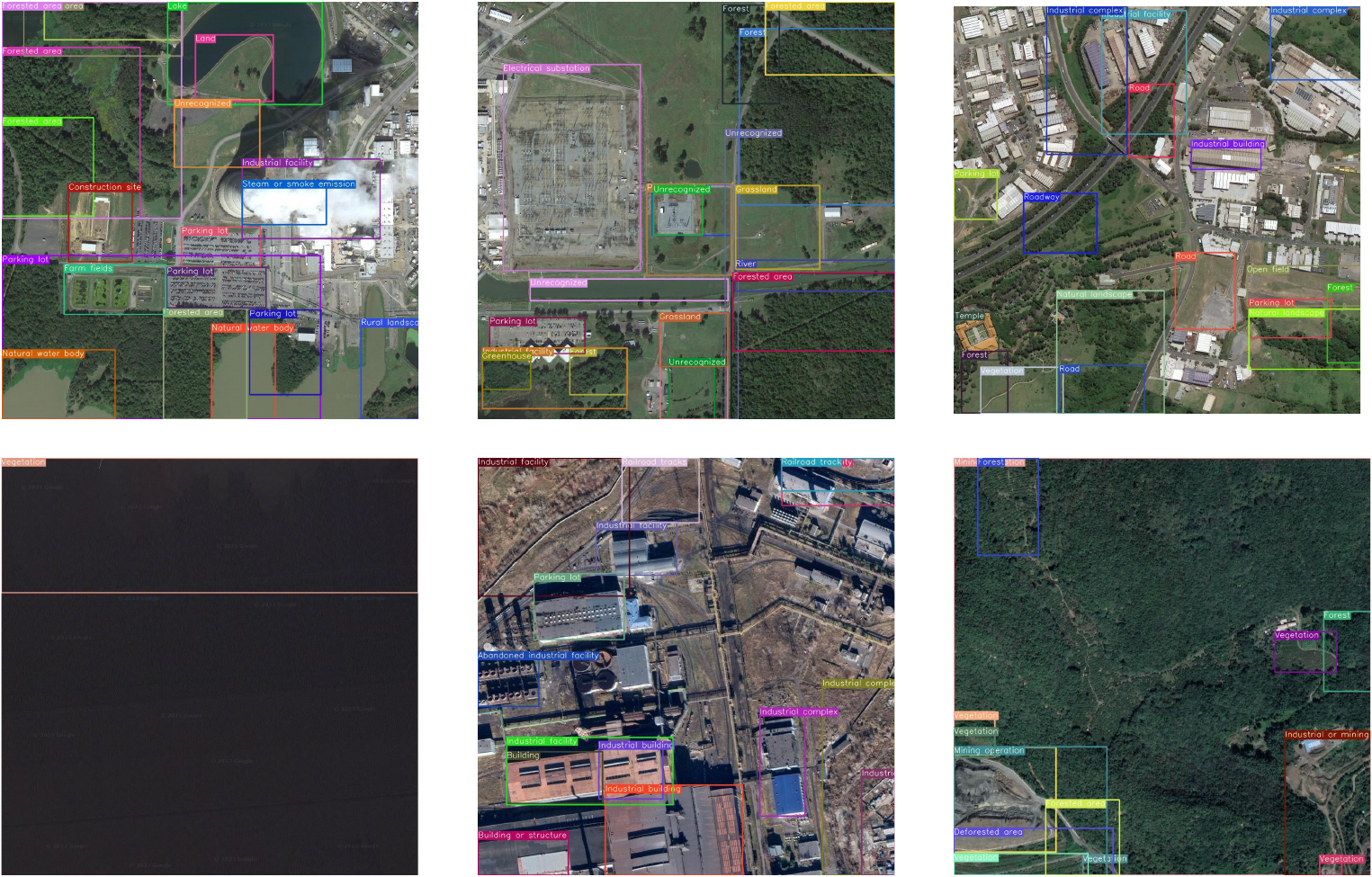}
    \caption{Raw data labelled by LAE-Label engine without rule-based filtering. The preliminary labelling results have already some accuracy and can provide a boost to the LAE model to understand Earth.}
    \label{fig:figA2}
\end{figure}

Figure \ref{fig:figA2} illustrates the raw image without rule-based filter annotation. The rule-based filtering approach employs the removal of monotonous images and the culling of some irrelevant and homogeneous categories.

\begin{table*}[t]
\tiny
\centering
\resizebox{\textwidth}{!}{
\begin{tabular}{p{1.0cm}|p{0.9cm}|p{5.8cm}|p{0.9cm}|p{5.5cm}}
\toprule
\multicolumn{1}{l|}{\textbf{Datasets}} & \textbf{Original Category Number} & \multicolumn{1}{l|}{\textbf{Original Categories}} & \textbf{Selected Category  Number} & \multicolumn{1}{l}{\textbf{Selected Categories}} \\ \midrule
DOTA & 18 & \begin{tabular}[c]{@{}l@{}}plane, ship, storage tank, baseball diamond, tennis court, \\ basketball court, ground track field, harbor, bridge, large \\ vehicle, small vehicle, helicopter, roundabout, soccer \\ ball field, swimming pool, container crane, airport, helipad\end{tabular} & 6 & \begin{tabular}[c]{@{}l@{}}helicopter, roundabout, soccer ball field, swimming pool, \\ container crane, helipad\end{tabular} \\ \hline
DIOR & 20 & \begin{tabular}[c]{@{}l@{}}airplane, airport, ground track field, harbor, baseball field, \\ overpass, basketball court, ship, bridge, stadium, storage \\ tank, tennis court, expressway service area, train station, \\ expressway toll station, vehicle, golf course, wind mill, chimney, dam\end{tabular} & 18 & \begin{tabular}[c]{@{}l@{}}airplane, airport, ground track field, harbor, baseball field, \\ overpass, basketball court, bridge, stadium, storage tank, \\ tennis court, expressway service area, train station, expressway \\ toll station, vehicle, golf course, wind mill, dam\end{tabular} \\ \hline
FAIR1M & 5(37) & \begin{tabular}[c]{@{}l@{}}airplane (Boeing 737, Boeing 777, Boeing 747, Boeing 787, \\ Airbus A320, Airbus A220, Airbus A330, Airbus A350, COMAC \\ C919, and COMAC ARJ21), ship (passenger ship, motorboat, \\ ﬁshing boat, tugboat, engineering ship, liquid cargo ship, \\ dry cargo ship, warship), vehicle (small car, bus, cargo truck, \\ dump truck, van, trailer, tractor, truck tractor, excavator), \\ court(basketball court, tennis court, football field, baseball \\ field), road (intersection, roundabout, bridge)\end{tabular} & 18 & \begin{tabular}[c]{@{}l@{}}passenger ship, motorboat, ﬁshing boat, tugboat, engineering \\ ship, liquid cargo ship, dry cargo ship, warship, small car, bus, \\ cargo truck, dump truck, van, trailer, tractor, truck tractor, \\ excavator, intersection\end{tabular} \\ \hline
Xview & 60 & \begin{tabular}[c]{@{}l@{}}Fixed-wing Aircraft, Small Aircraft, Cargo Plane, Helicopter, \\ Passenger Vehicle, Small Car, Bus, Pickup Truck, Utility Truck, \\ Truck, Cargo Truck, Truck w/Box, Truck Tractor, Trailer, Truck \\ w/Flatbed, Truck w/Liquid, Crane Truck, Railway Vehicle, \\ Passenger Car, Cargo Car, Flat Car, Tank car, Locomotive, \\ Maritime Vessel, Motorboat, Sailboat, Tugboat, Barge, Fishing \\ Vessel, Ferry, Yacht, Container Ship, Oil Tanker, Engineering \\ Vehicle, Tower crane, Container Crane, Reach Stacker, Straddle \\ Carrier, Mobile Crane, Dump Truck, Haul Truck, Scraper/Tractor, \\ Front loader/Bulldozer, Excavator, Cement Mixer, Ground Grader, \\ Hut/Tent, Shed, Building, Aircraft Hangar, Damaged Building, \\ Facility, Construction Site, Vehicle Lot, Helipad, Storage Tank, \\ Shipping container lot, Shipping Container, Pylon, Tower\end{tabular} & 34 & \begin{tabular}[c]{@{}l@{}}Fixed-wing Aircraft, Small Aircraft, Cargo Plane, Pickup Truck, \\ Utility Truck, Passenger Car, Cargo Car, Flat Car, Locomotive, \\ Sailboat, Barge, Ferry, Yacht, Oil Tanker, Engineering Vehicle, \\ Tower crane, Reach Stacker, Straddle Carrier, Mobile Crane, \\ Haul Truck, Front loader/Bulldozer, Cement Mixer, Ground Grader, \\ Hut/Tent, Shed, Building, Aircraft Hangar, Damaged Building, \\ Facility, Construction Site, Shipping container lot Shipping \\ Container, Pylon, Tower\end{tabular} \\ \hline
Condensing-Tower & 4 & \begin{tabular}[c]{@{}l@{}}working chimney, unworking chimney, working condensing \\ tower, unworking condensing tower\end{tabular} & 4 & \begin{tabular}[c]{@{}l@{}}working chimney, unworking chimney, working condensing \\ tower, unworking condensing tower\end{tabular} \\ \bottomrule
\end{tabular}}
\caption{LAE-80C is sampled from the validation set of multiple remote sensing object detection datasets to filter the categories that are as semantically non-overlapping as possible.}
\label{tab:tableA2}
\end{table*}

\subsubsection{LAE-1M Word Cloud.} 
\begin{figure}[t]
    \centering
    \includegraphics[width=\linewidth]{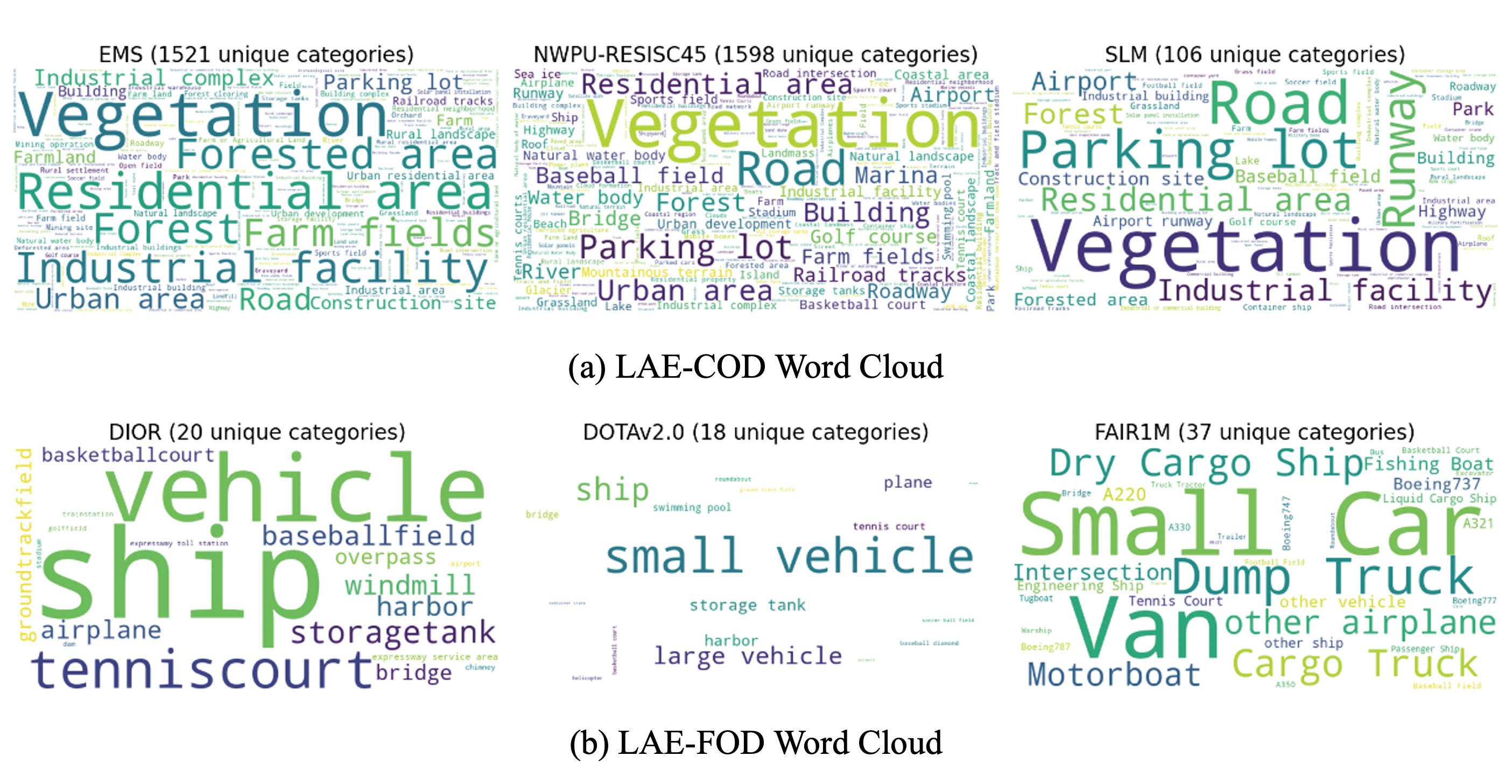}
    \caption{LAE-1M Word Cloud (Part). LAE-COD carries dense categories, while LAE-FOD categories are sparse. \textcolor{black}{Although dense categories include unbounded ones like roads, vegetation, and water, sparse categories are better understood by learning this part.}}
    \label{fig:figAA3}
\end{figure}

Figure~\ref{fig:figAA3} presents a word cloud of a subset of LAE-1M, illustrating that LAE-COD encompasses a richer variety of semantic categories than the LAE-FOD. This diversity aids in improving open-vocabulary modeling for the remote sensing community.

\subsubsection{LAE-80C Benchmark.} In order to expand the number of classes in the DIOR \cite{li2020object} and DOTAv2.0 \cite{Xia_2018_CVPR} datasets, we have introduced additional classes selected from various sub-datasets’ test sets. This was done to minimize class duplication, as depicted in Table \ref{tab:tableA2}. We prioritized preserving the classes from the high-quality DIOR dataset and then added the categories not available in DIOR from the DOTAv2.0 dataset. For more detailed category data, we included categories from FAIR1M \cite{sun2022fair1m} and Xview \cite{lam2018xview} datasets. Additionally, we incorporated the categories of different states in the Condensing-Tower \cite{zhang2019deep} dataset. In the end, we combined these categories to create a benchmark with 80 categories.

\section{More Implementation Details}
\subsubsection{LAE-Label Engine.} For the \textbf{SAM} model, we use a version of SAM-ViT-H, along with a series of hyperparameters in the execution, where we take 32 points randomly for each image. Then, the threshold for IOU is set to 0.86, the threshold for score is set to 0.92, and the downsampling factor is set to 2. For images cropped to 1024 resolution size, we take the top ten RoIs with the largest area, while for small resolution images, the top five RoIs with the largest area are used. For the \textbf{InternVL} model, we use the InternVL-1.5 version, which performs parallel inference on eight A100s and completes inference on a single image in about 1.5  on average.

\subsubsection{LAE-DINO Model.} We conducted all pre-training experiments on four A100 GPUs. The prediction heads' categories are also set to 1600 for open-vocabulary pre-training. During training, key parameters are carefully set: the length of the dynamic vocabulary number in the DVC module, i.e., $N_{\mathcal{DV}}$ is set to 60, the number of layers $l$ for MDSA and FFN is set to 7, and the hyper-parameters $\alpha$ and $\beta$ of the loss function are set to 1 and 10, respectively. The open-vocabulary pre-training of LAE-DINO lasts approximately 180$K$ steps, spanning about 48 GPU hours with a batch size of 2 per GPU. When fine-tuning the DIOR and DOTAv2.0 datasets, we set the size of the prediction heads' categories to 20 and 18, respectively. The visual and textual backbone used are Swin-T and BERT, respectively. We use AdamW \cite{kingma2014adam} as the optimiser, with a learning rate $10^{-4}$ and a weight decay factor set to $10^{-4}$. Followed the GroudingDINO \cite{liu2024grounding}, $\lambda_{L_1}$ and $\lambda_{GIoU}$ are set 5.0 and 2.0, respectively.

\section{More Results}
\subsection{Few-shot Detection.} Table \ref{tab:tableA7} demonstrates the results of the few-shot detection on the HRRSD test set. The table result further demonstrates that our method has a better recognition of both base classes and unseen classes.
\begin{table}[t]
\tiny
\centering
\resizebox{\linewidth}{!}{%
\begin{tabular}{l|l|ccc}
\toprule
\multirow{2}{*}{\textbf{}} & \multirow{2}{*}{\textbf{Method}} & \multicolumn{3}{c}{\textbf{HRRSD}} \\ &  & $mAP_{base}$ & $mAP_{novel}$ & $mAP_{all}$ \\ \hline
\multicolumn{1}{c|}{\multirow{3}{*}{1-shot}}  & GLIP & 39.3 & 1.4 & 30.6  \\
\multicolumn{1}{c|}{}& GroudingDINO & 37.8 & 3.3 & 29.8  \\
\multicolumn{1}{c|}{}& \cellcolor{mygray}LAE-DINO & \cellcolor{mygray}40.2 & \cellcolor{mygray}5.0 & \cellcolor{mygray}32.1  \\ \hline
\multicolumn{1}{c|}{\multirow{3}{*}{5-shot}}  & GLIP & 38.5 & 10.5 & 32.0  \\
\multicolumn{1}{c|}{}  & GroudingDINO & 43.0 & 11.2 & 35.7  \\
\multicolumn{1}{c|}{}   & \cellcolor{mygray}LAE-DINO & \cellcolor{mygray}42.5 & \cellcolor{mygray}13.7 & \cellcolor{mygray}35.9  \\ \hline
\multicolumn{1}{c|}{\multirow{3}{*}{10-shot}}  & GLIP & 47.6 & 13.8 & 39.8  \\
\multicolumn{1}{c|}{}  & GroudingDINO & 48.0 & 15.8 & 40.6  \\
\multicolumn{1}{c|}{}   & \cellcolor{mygray}LAE-DINO & \cellcolor{mygray}50.4 & \cellcolor{mygray}15.8 & \cellcolor{mygray}42.4  \\ \bottomrule
\end{tabular}}
\caption{The few-shot detection results on the HRRSD test set with the ten base classes appearing in the LAE-1M dataset and three novel classes that do not, i.e., \textit{T junction}, \textit{crossroad}, and \textit{parking lot}.}
\label{tab:tableA7}
\end{table}

\subsubsection{LAE-COD Quality Evaluation.} 
\begin{figure}[t]
    \centering
    \includegraphics[width=\linewidth]{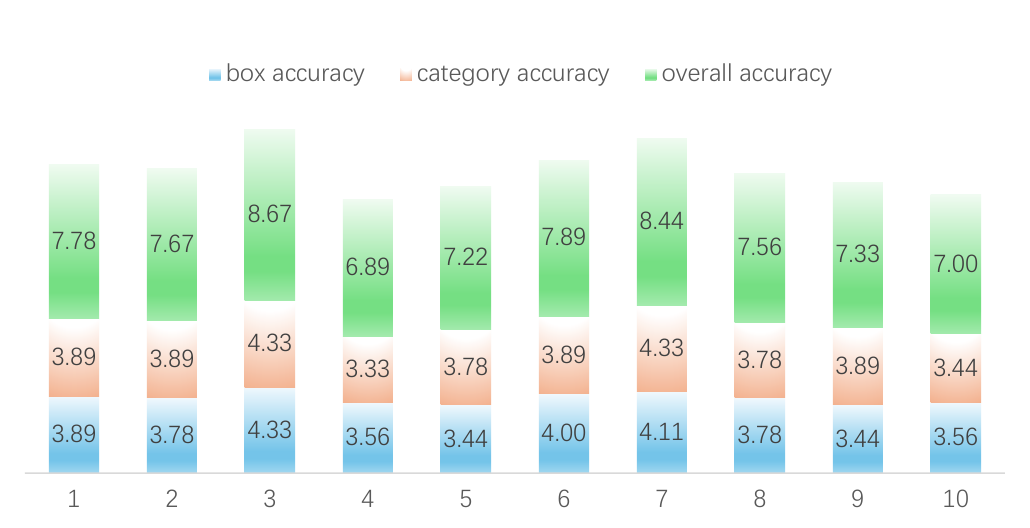}
    \caption{The result of a randomly ten selected sample is scored out of 5 for the box accuracy and the category accuracy and out of ten for the overall accuracy.}
    \label{fig:figA9}
\end{figure}

To further verify the LAE-Label engine's annotation quality, we randomly sampled ten image annotations from LAE-COD. Then, we provided some users a quality score (1-5 points) from \textit{category accuracy} and \textit{box accuracy}, respectively. The results of the survey are shown in Figure \ref{fig:figA9}. This survey shows that the overall quality of annotation is good, and the quality of classification is generally better than the quality of border. Most adopt a positive attitude towards the accuracy and completeness of the labeling but still need to pay attention to the further improvement of the border quality.

\subsubsection{LAE-1M Reanalysis.} To explore how LAE-COD and LAE-FOD of LAE-1M work, we set up two sets of comparison experiments on our LAE-DINO as shown in Table \ref{tab:tableAA5}. We find that the detection of base classes in LAE-FOD can be improved by adding additional LAE-COD for pre-training, where the $mAP$ of DOTAv2.0 test set can be improved by 2.3\%. This also implies the feasibility of our LAE-Label to help interpret common categories of remote sensing imagery.
\begin{table}[h]
\tiny
\centering
\resizebox{\linewidth}{!}{%
\begin{tabular}{l|c|ccc}
\toprule
\multirow{2}{*}{\textbf{Method}} & \multirow{2}{*}{\textbf{Pre-Training Data}} & \multicolumn{1}{c}{\textbf{DIOR}} & \multicolumn{1}{c}{\textbf{DOTAv2.0}} & \multicolumn{1}{c}{\textbf{LAE-80C}} \\ &  & $AP_{50}$ & $mAP$ & $mAP$\\ \hline
LAE-DINO   & LAE-FOD &  84.1 & 44.5 & 19.1 \\
\rowcolor{mygray}LAE-DINO  & LAE-FOD$+$LAE-COD & 85.5(+1.4) & 46.8(+2.3) & 20.2(+1.1)  \\
\bottomrule
\end{tabular}}
\caption{The open-set detection results with different pre-training data.}
\label{tab:tableAA5}
\end{table}

\begin{figure}[t]
    \centering
    \includegraphics[width=\linewidth]{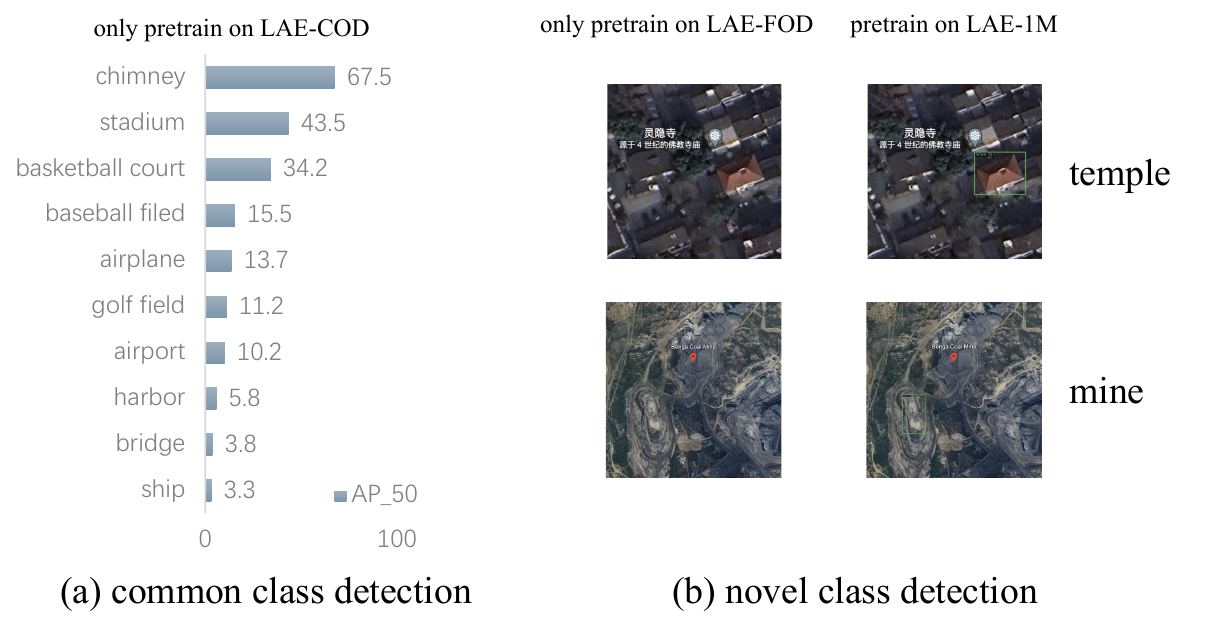}
    \caption{The role of LAE-COD dataset in common class detection and novel class detection. The source of the image to be detected is from Google Earth.}
    \label{fig:figA7}
\end{figure}

The LAE-Label engine identifies common categories in remote sensing object detection and enhances the detection of some novel categories.
Figure \ref{fig:figA7} shows the detection results of the LAE-COD dataset on the DIOR common classes and some uncommon novel classes. Figure \ref{fig:figA7}(a) represents the zero-shot detection results on the DIOR test set only after LAE-COD training, where $AP_{50}$ of the category is the top 10 category. Figure \ref{fig:figA7}(b) shows that the detection results of some rare classes are pre-trained with or without LAE-COD, which cannot be quantitatively evaluated because there is no corresponding data set, and the data was obtained from Google Earth. We can find that this semi-automated labeling approach LAE-Label significantly improves the shortcomings of existing remote sensing object detection datasets. Additionally, including more categories offers a feasible solution for the LAE task.

\subsubsection{VisGT Reanalysis.}
\begin{figure}[t]
    \centering
    \includegraphics[width=0.9\linewidth]{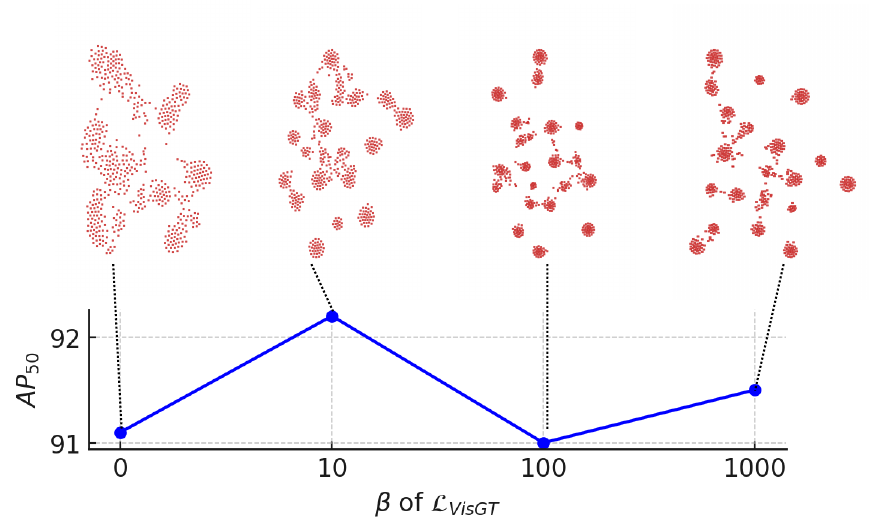}
    \caption{The detection results on the DIOR test set with different $\beta$ values of $\mathcal{L}_{VisGT}$, which is visualised the visual interpolation of VisGT at different $\beta$, using the t-SNE \cite{van2008visualizing} method.}
    \label{fig:figA3}
\end{figure}

Inspired by some remote sensing image-text retrieval work\cite{pan2023reducing}, we visualise the distribution of image features for VisGT. Figure \ref{fig:figA3} shows the detection results on the DIOR test set with different $\beta$ values of $\mathcal{L}_{VisGT}$. We also visualized the visual interpolation of VisGT at different $\beta$, using the t-SNE \cite{van2008visualizing} method. The clustered red squares represent the visual features of the image as a whole in semantic space. It can be observed that imposing constraints on VisGT will be better than the prompts without constraints, but over-imposing constraints can also affect the detection results. VisGT works best, especially when $\beta$ is around 10. VisGT can control $\beta$ of $\mathcal{L}_{VisGT}$ to achieve engagement with visual-guided textual prompts, supplementing insufficient textual information to influence the final detection results. In complex remote sensing scenes, VisGT can first retrieve the approximate scene by mapping the visual features into the semantic space and then perform the detection.

\subsubsection{Visualization.}
\begin{figure}[t]
    \centering
    \includegraphics[width=0.9\linewidth]{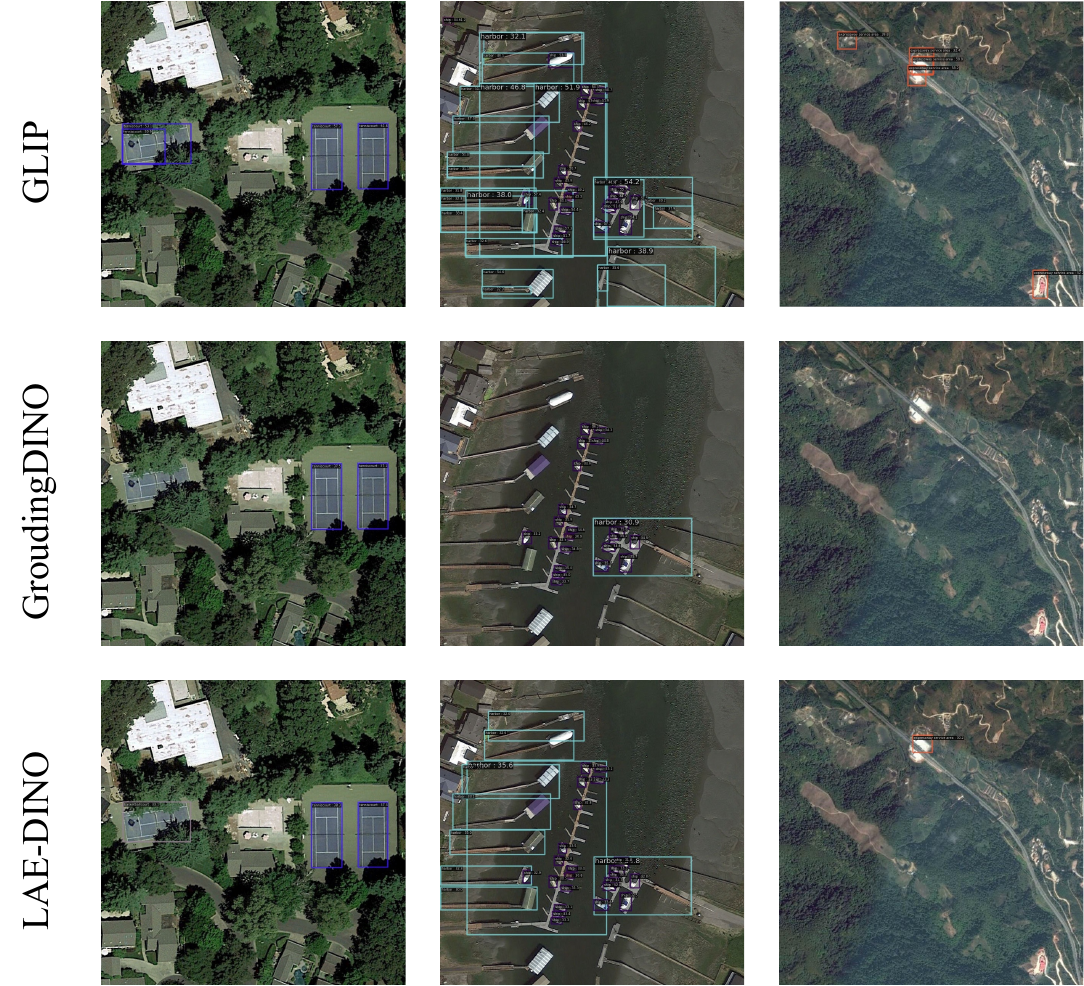}
    \caption{Visualization of GLIP, GroudingDINO and LAE-DINO in open-set detection, both pre-trained on LAE-1M dataset.}
    \label{fig:figA1}
\end{figure}

We visualize open-set detection for GLIP \cite{li2022grounded}, GroundingDINO \cite{liu2024grounding}, and LAE-DINO, as shown in Figure \ref{fig:figA1}. We select a few common categories, such as \textit{tennis court}, \textit{ship} and \textit{harbor}, and \textit{expressway service area}. Through observation, it is found that GLIP results are biased towards finding more that may be similar, while GroundingDINO finds it as accurate as possible, even if it misses some. Our LAE-DINO detection results are superior to the other two methods. It also side-steps the effectiveness of retrieving scenarios before the detection approach.

\end{document}